\documentclass[letterpaper, 10 pt, conference]{ieeeconf}  

\IEEEoverridecommandlockouts                              

\usepackage{graphics} 
\usepackage{epsfig} 
\usepackage{times} 
\usepackage{amsmath} 
\usepackage{amsthm}
\usepackage{amssymb}
\usepackage{tikz}
\usetikzlibrary{graphs}
\usepackage[colorlinks=false, urlcolor=blue, pdfborder={0 0 0}]{hyperref}
\usepackage{enumerate}
\usepackage{color}
\usepackage{bbm}

\usepackage{algorithm}
\usepackage{cite}
\usepackage{comment}

\usepackage{relsize}
\usepackage{amsfonts}
\usepackage{subcaption}
\usepackage{amsthm}
\usepackage{multirow}
\usepackage{multicol}
\usepackage{comment}
\newtheorem{theorem}{Theorem}

\newtheorem{lemma}[theorem]{Lemma}
\theoremstyle{definition}
\newtheorem{definition}{Definition}

\DeclareMathOperator*{\argmin}{arg\,min}
\usepackage{array}
\usepackage[noend]{algpseudocode}

\usepackage{xcolor}

\usepackage{newfloat}
\usepackage{listings}
\DeclareCaptionStyle{ruled}{labelfont=normalfont,labelsep=colon,strut=off} 
\floatstyle{ruled}
\newfloat{listing}{tb}{lst}{}
\floatname{listing}{Listing}
\title{Learning Adaptive Safety for Multi-Agent Systems}

\author{
Luigi Berducci$^{1a}$, 
Shuo Yang$^{2}$, 
Rahul Mangharam$^{2}$, 
Radu Grosu$^{1}$
\thanks{$^{1}$Institute of Computer Engineering, TU Wien} 
\thanks{$^{2}$Dept. of Electrical and Systems Engineering, University of Pennsylvania}
\thanks{$^{a}$Correspondance to:
{\tt\small luigi.berducci@tuwien.ac.at}
}%
}

\begin{document}

\maketitle
\thispagestyle{empty}
\pagestyle{empty}

\begin{abstract}
Ensuring safety in dynamic multi-agent systems is challenging due to limited information about the other agents. 
Control Barrier Functions (CBFs) are showing promise for safety assurance
but current methods make strong assumptions about other agents and often rely on manual tuning to balance safety, feasibility, and performance.
In this work, 
we delve into the problem of adaptive safe learning 
for multi-agent systems with CBF.
We show how emergent behavior can be profoundly influenced by the CBF configuration, highlighting the necessity for a responsive and dynamic approach to CBF design.
We present ASRL, a novel adaptive safe RL framework, to fully automate the optimization of policy and CBF coefficients,
to enhance safety and long-term performance through reinforcement learning.
By directly interacting with the other agents, ASRL learns to cope with diverse agent behaviours and maintains the cost violations below a desired limit. 
We evaluate ASRL in a multi-robot system and a competitive multi-agent racing scenario,
against learning-based and control-theoretic approaches.
We empirically demonstrate the efficacy and flexibility of ASRL, and 
assess generalization and scalability to out-of-distribution scenarios.
Code and supplementary material are public online\footnote{All code and supplementary material: \url{https://github.com/luigiberducci/learning_adaptive_safety}}.
\end{abstract}

\section{Introduction}
Safety is an outstanding concern in the design of learning algorithms, 
especially 
%
for safety-critical applications.  
Control barrier functions (CBFs) have emerged in this context as a very powerful formal approach to ensuring safety~\cite{ames2016control, dawson2023safe, yang2022differentiable}. Moreover, the integration of CBFs in re\-in\-for\-ce\-ment learning (RL) holds a huge potential for safe exploration~\cite{cheng2019end, emam2022safe, choi2020reinforcement, wang2023multi, li2019temporal}.

However, the success of CBFs in RL is often confined to simple settings, such as single-agents or cooperative multi-agents with very limited interaction. 
This is because in multi-agent scenarios, the intricate interplay among agents poses unique challenges to the design of the CBFs and their associated \emph{extended class-$\mathcal{K}_{\infty}$ functions}.
These functions, in the following abbreviated as \emph{class-$\mathcal{K}$ functions}, control the rate with which the agent can approach the safe-set boundary.


\begin{figure}[ht!]
    \centering
    \includegraphics[width=1.0\columnwidth]{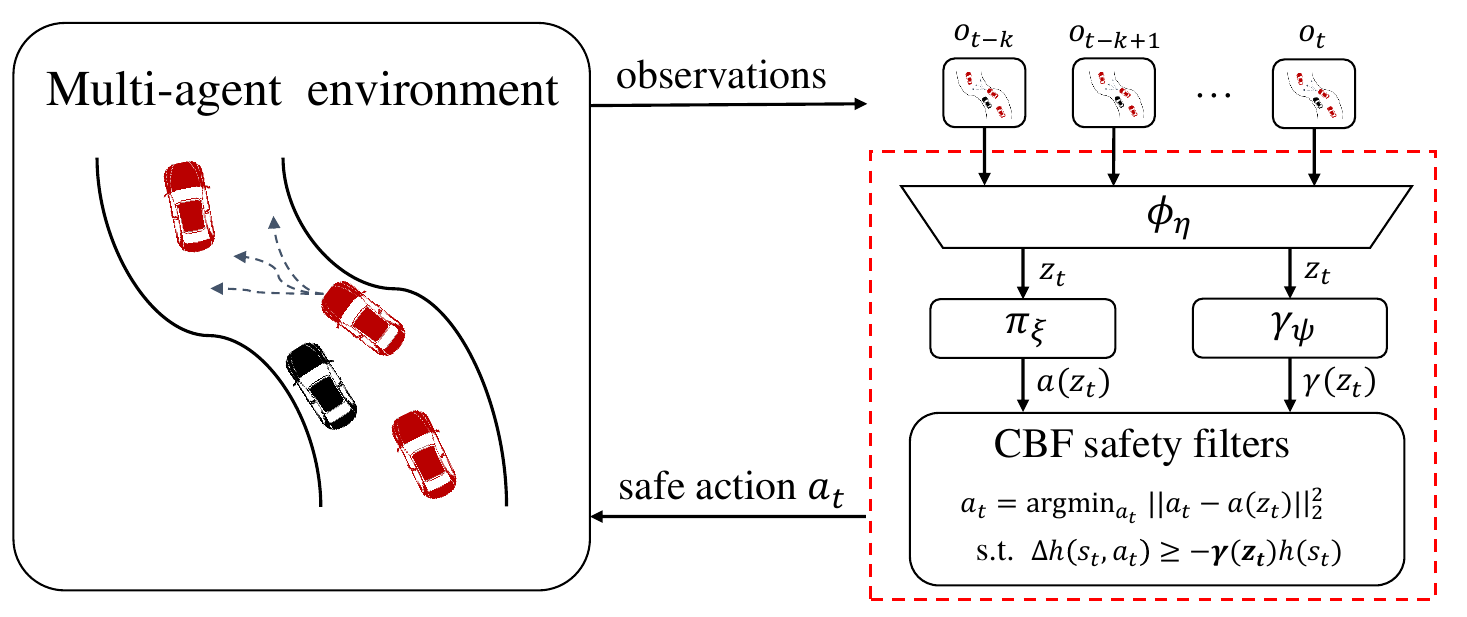}
    \caption{
    The proposed hierarchical adaptive framework for multi-agent systems,
    where a policy $\pi_{\xi}$ and a safety module $\gamma_{\psi}$ are jointly optimized for safe and adaptive interaction.}
    \label{fig:framework}
    \vspace*{-3ex}
\end{figure}

While manual tuning of the class-$\mathcal{K}$ functions is feasible in simple tasks, it becomes challenging in multi-agent environments due to the unpredictable effects of small changes in their parameters. 
The richness of interactions and limited information about other agents' policies make it difficult to trade-off safety and long-term performance by adjusting just a few parameters.
%
Previous work on Adaptive CBFs have primarily concentrated on enhancing the feasibility of the optimization problem, by introducing time-varying coefficients within the CBF condition \cite{xiao2021adaptive}. 
However, it's worth noting that these approaches have been predominantly applied in single-agent or cooperative environments, 
often relying on the availability of substantial historical data for optimization \cite{qin2021learning, xiao2023barriernet}.
These methods face two key challenges:
\begin{itemize}
    \item \textit{Overlooking long-term objectives}: 
    the narrow focus on feasibility and short-term performance within the adaptive CBF framework,
    struggles in capturing long-term objectives and potentially leads to sub-par solutions.
    \item \textit{Prior-data scarcity}: 
    the assumption that historical data is available does not always hold, 
    especially in non-cooperative 
    settings 
    where other agents may be reluctant to reveal their strategies, 
    thereby hindering the achievement of sufficient coverage of diverse strategies.
\end{itemize}

To address these two challenges in interactive multi-agent environments, 
we propose a novel approach based on RL and adaptive CBFs. 
In order to account for the lack of knowledge with regard to the other agents' strategies, we exploit direct interactions with these agents, to uncover their intentions.

\vspace{1mm}
\noindent Our \textbf{main contributions} in this paper are the following:
\begin{enumerate}    
\item\emph{An adaptive safe-RL framework (ASRL)}, where a low-level CBF-controller ensures safety and a high-level one optimizes policy and state-dependent CBF coefficients.
%
\item \emph{A model-free learning approach}, which is based on RL for efficient adaptation to different agents and scenarios through direct interaction with these agents.
\item\emph{A comprehensive evaluation of ASRL} in multi-agent environments, in order to assess the adaptation to different types of agents and degrees of cooperation.    
\end{enumerate}

As shown in Figure~\ref{fig:framework}, by combining a model-based low-level control layer with model-free RL, ASRL enhances adaptiveness to diverse behaviors exhibited by other agents,
relieving the engineers from the burden of manual tuning the CBF,
in favour of a systematic approach to optimize general long-term objectives and trade-off safety and performance.


\subsubsection{Motivating Example}{
\label{section:motivating_example}
Consider a navigation task where multiple robots have distinct starting positions and specific goals.
The controlled \textit{ego} robot needs to reach its goal while avoiding collisions with other robots, without any knowledge of their parameterization.
To ensure safe navigation, we equip the ego robot with a CBF, acting as a protective safety shield.
However, the emergent behavior of the ego robot can vary significantly by adjusting the coefficient $\gamma$ of the CBF class-$\mathcal{K}$ function, as shown in Figure \ref{fig:motivating_example} (left). 
In some simulations, the ego fails to reach its goal due to cautious maneuvers dictated by the CBF condition, while the same controller successfully completes the task in other configurations.

Two factors contribute to this:
(1) the ego robot adapts its maneuvers based on the CBF condition, varying assertiveness levels;
(2) other robots react to the ego's actions, leading to configurations that can aid or hinder task completion.
%
%
Figure \ref{fig:motivating_example} (right) supports this hypothesis 
by showing how diverse CBF coefficients influence the long-term performance of the ego agent,
measured by success and collision rates.
The optimal coefficients depend on scenario-specific characteristics, 
such as the number of agents and their parameters.
This underscores the importance of an adaptive approach,
which will be detailed in the following sections of this work.


\begin{figure*}[h!]
    \centering
    \includegraphics[width=\textwidth]{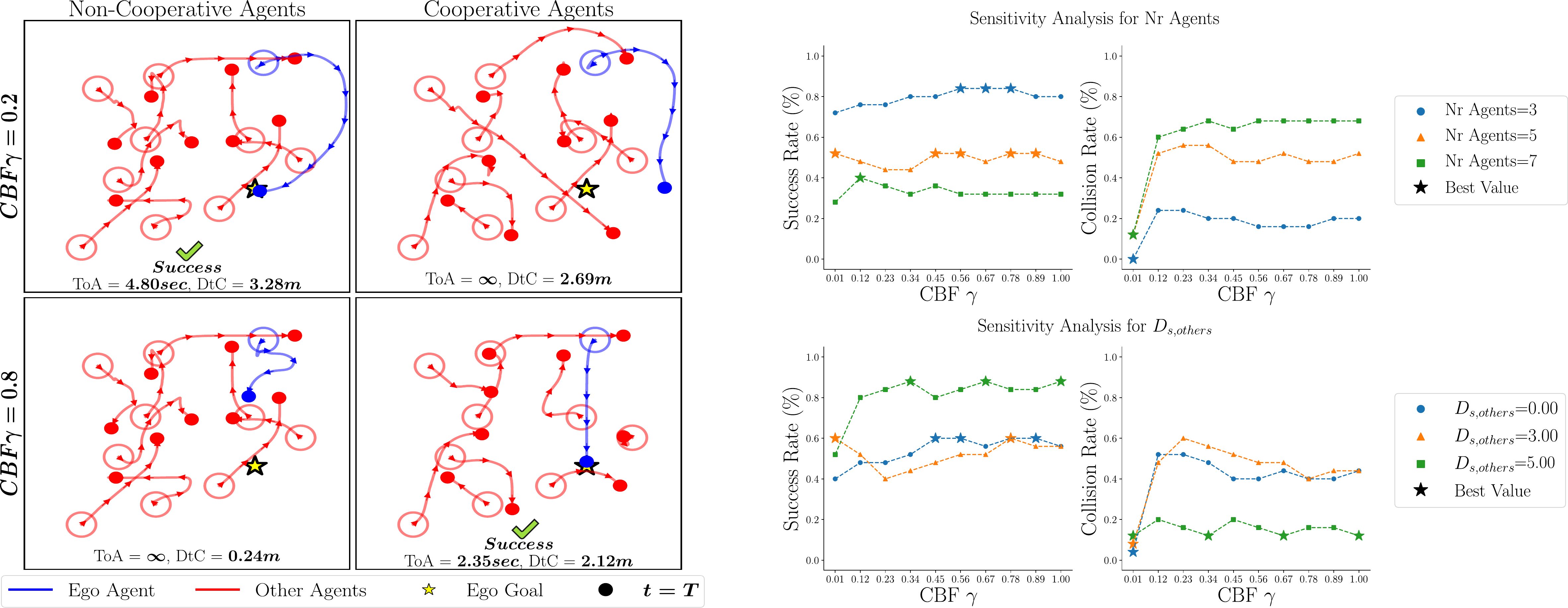}
    \caption{
    \textbf{Left: } Simulations with non-cooperative (left) and cooperative agents (right) under different CBF coefficients $\gamma$. 
    Time of Arrival (\textit{ToA}) and Minimum Distance to Collision (\textit{DtC}) are reported for the ego agent (\textit{blue}).
    The long-term effects diverge based on the coefficients, showing the importance of adaptation.
    \textbf{Right:} Average performance of CBF coefficients under different number of agents (\textit{top}) and safety distances $D_{s,others}$ (\textit{bottom}) of other agents' controllers ($n=25$).
    }
    \label{fig:motivating_example}
    \vspace{-4ex}
\end{figure*}

}

\section{Background}

Consider the stochastic game 
$(\mathcal{I}, S, \bar{A}, f, r, \rho_0, T, \alpha)$, 
where $\mathcal{I}=\{1, 2, \cdots, q\}$ denotes the set of $q$ agents,
$S$ and $\bar{A}$ are the set of states and joint actions,
$f : S\times A \rightarrow S$ is the deterministic transition function,
$r : S \times A \times S \rightarrow \mathbb{R}$ is the reward function,
$\rho_0$ represents the distribution of initial conditions,
$T \in \mathbb{N}$ denotes the time horizon,
$\alpha \in [0, 1]$ is the discount factor, to avoid confusion with the $\gamma$ adopted in CBF.
At timestep $t$, 
each agent $i$ picks an action $a^i_t$ according to its policy $\pi_i$ and 
the system state evolves according to the joint action $a_t = \times_{i \in \mathcal{I}}~a^i_t$ by discrete-time dynamics
\begin{equation}
s_{t+1} = f(s_t, a_t).
\label{eq:transition_dynamics}
\end{equation}
We consider the problem of finding an optimal policy for the \textit{ego} agent $\pi_1$,
assuming the non-controlled agents $\pi_i, i > 1,$ have unobservable parameters
distributed according to $\rho_0$.

In the following, we denote the ego with $\pi$ and formulate the problem as a 
constrained partially observable Markov decision process (CPOMDP)
$(S, A, \Omega, O, f, r, h, \rho_0, T, \alpha)$
where 
the actions $A$ refer to $\pi$,
the observations $\Omega$ consists of 
the observable states without the other agents' parameters obtained by $O: S \rightarrow \Omega$,
$h$ is a continuously differentiable function delimiting the safe set of states for the ego agent.
We define the safe set, $\mathcal{C}$, by the superlevel set of $h$:
\begin{align}\label{eq:safe_set}
 \mathcal{C}&=\{s\in S:h(s)\ge 0\}.
\end{align}

\begin{definition}(Forward invariance and safety)
The set $\mathcal{C}$ is \emph{forward invariant} if for every $s_0\in \mathcal{C}$, $s_t\in \mathcal{C}$ holds for all $t$. If $\mathcal{C}$ is forward invariant, we say the system~\eqref{eq:transition_dynamics} is safe. 
\end{definition}

\begin{definition}(CBF~\cite{zeng2021safety})
Given a set $\mathcal{C} \subset \mathbb{R}^n$ defined by \eqref{eq:safe_set}, the continuously differentiable function $h: \mathbb{R}^n \rightarrow \mathbb{R}$ is a \textit{discrete-time control barrier function} (CBF) for dynamical system (\ref{eq:transition_dynamics}) if there exists $\gamma \in [0,1]$  such that for all $s_t \in \mathcal{C}$, 
\begin{equation}
\sup_{a_t \in A} \Big[ h\Big(f(s_t, a_t)\Big) + (\gamma - 1) h(s_t) \Big] \geq 0.
\label{eq:discrete_CBF}
\end{equation}
\end{definition}

Note that the parameter $\gamma$ influences the conservativness of agent's behaviour: it will be less conservative (i.e., approaching the safe boundary) as $\gamma$ goes to 1.
However, $\gamma$ is fixed in the above vanilla CBF definition, which implies the fixed degree of conservativness.
To overcome this limitation, 
we introduce the adaptive version of discrete-time CBF.
\begin{definition}(Adaptive control barrier function)
Given a set $\mathcal{C} \subset \mathbb{R}^n$ defined by (\ref{eq:safe_set}), the continuously differentiable function $h: \mathbb{R}^n \rightarrow \mathbb{R}$ is a \textit{discrete-time adaptive control barrier function} (Adaptive CBF) for dynamical system (\ref{eq:transition_dynamics}) if for all $s_t \in \mathcal{C}$, there exists $\gamma(s_t) \in [0,1]$ such that  
\begin{equation}
\sup_{a_t \in A} \Big[ h\Big(f(s_t, a_t)\Big) + (\gamma(s_t) - 1) h(s_t) \Big] \geq 0.
\label{eq:adaptive_discrete_CBF}
\end{equation}
\end{definition} 

The Adaptive CBF differs from \eqref{eq:discrete_CBF}
by the state-dependent function $\gamma: S \rightarrow [0,1]$.
We now demonstrate that state-dependent coefficients do not hinder the safety guarantees of CBF.
For any potentially unsafe nominal action $a^{nom}_t$, 
we can obtain a safe action solving the quadratic program (QP):
\begin{equation}
\begin{aligned}
a_t = ~ & \underset{a_t\in A}{\text{argmin}}
& & \| a_t - a^{nom}_{t} \|^2_2\\
& ~ \text{s.t.}
& &   h(f(s_t, a_t)) + (\gamma(s_t) - 1) h(s_t) \geq 0.
\end{aligned}
\label{eq:qp}
\end{equation}
%
%
\begin{lemma}
	For dynamical system (\ref{eq:transition_dynamics}), if the QP problem in (\ref{eq:qp}) is feasible for all $s \in \mathcal{C}$, then the controller derived from (\ref{eq:qp}) renders set $\mathcal{C}$ forward invariant, i.e., safety is preserved. 
	\label{lemma:lemma_barrier}
    \vspace{-3ex}
\end{lemma}
\begin{proof}
    For any initial state $s_0\in\mathcal{C}$, we can derive that:
    \begin{align}
        h(s_t)&=h(f(s_{t-1}, a_{t-1})) \ge (1 - \gamma(s_{t-1}))\, h(s_{t-1})\nonumber\\
        &\ge (1 - \gamma(s_{t-1}))\, (1 - \gamma(s_{t-2})) \cdot h(s_{t-2})\nonumber\\
        &\cdots\nonumber\\
        &\ge \prod_{i=0}^{t-1}(1 - \gamma(s_{i}))\, h(s_{0})\ge 0
    \end{align}
    which implies that safety is preserved at any time $t$.
\end{proof}
Practical CBF usability in multi-agent systems comes with a few important remarks.
Problem (\ref{eq:qp}) is tipically non-convex,
and prior work  focused on linear CBFs and convex formulations, 
for improved efficiency and optimal solutions~\cite{cheng2019end}. 

In contrast, ASRL does not assume such structures 
or the ability to optimally solve the QP. 
Instead, ASRL adaptiveness uses class-$\mathcal{K}$ functions, 
allowing users to provide any CBF.  
Moreover, 
handling multiple CBF constraints and input limits introduces feasibility issues
\cite{tan2022compatibility, black2023consolidated}. 
This is further exacerbated in multi-agent systems, 
where limited information about the other agents preclude achieving full safety guarantees.
These uncertainties can be quantified and integrated \textit{explicitly} 
in robust CBF formulations, with probabilistic guarantees \cite{cheng2020safe}.
In this work, 
the uncertainty is \textit{implicitly} learned in the adaptive model and the safety relaxed into a chance constraint, as formulated in next section.


\section{Problem Statement}
\label{section:problem_statement}

We consider the problem of learning a policy $\pi_\theta$
with parameters $\theta$ 
for an agent operating in a multi-agent en\-vi\-ron\-ment,
with partial observability of the other agents' parameters.
We formulate it as a CPOMDP with a cost function $h$
and aim to find a solution that keeps 
the occurrence of safety violations below a desired level of tolerance $d \in [0, 1]$.

Formally, the policy optimization problem is defined as:
\vspace{-2mm}
\begin{align}\label{eq:policy_optimization}
\max_{\theta}&~\mathcal{J}_R(\theta) = 
\mathbb{E}_{\tau \sim \pi_\theta} \big[ \sum_t^H \alpha^t \, r(s_t, a_t, s_{t+1}) \big] \\
\text{s.t. } &\mathcal{J}_C(\theta) = \mathbb{P}_{\tau \sim \pi_\theta}(h(s_t) < 0) \leq d \nonumber
\end{align}
%
%
%
where the trajectory $\tau = (s_0, a_0, s_1, a_1, ...)$ 
results from the
interaction 
of $\pi$ with the agents $\pi_{i > 1}$, 
given the initial distribution $s_0, \pi_{i>1} \sim \rho_0$,
and dynamics $s_{t+1} = f(s_t, a_t)$.

\section{Adaptive Safe Reinforcement Learning}



We present ASRL, the main contribution of this work,
an adaptive framework for multi-agent systems, 
which 
combines low-level model-based control and
model-free RL in a hierarchical fashion,
and the associated optimization algorithm.

\vspace{.5ex}
\noindent \textbf{Hierarchical Model Architecture.}
We structure the autonomous agent $\pi$ into a high-level model,
which drives the system towards the desired goal and provides an adaptive class-$\mathcal{K}$ function,
and a low-level layer,
which enforces the system safety using the barrier function $h$, the actions, and the coefficients from the high-level model.

To address partial observability of other agents, 
we adopt a novel multi-head actor with the following components:
\vspace{-3ex}

\begin{align}
    &\text{Representation model:} &z_t = \phi_\eta(o_{t-k:t}) \\
    &\text{Policy model:} &a_t \sim \pi_\xi(~\cdot~|~z_t) \\
    &\text{Safety model:} &\gamma_t \sim \gamma_\psi(~\cdot~|~z_t)
\end{align}

It consists of a representation model $\phi_\eta$ which encodes the past $k$ observations
$o_{t-k:t}$ into an embedding $z$,
a policy head $\pi_\xi$ which produces action $a_t$, and 
an adaptive-safety head $\gamma_\psi$ which outputs the CBF coefficient $\gamma_t$.
Our multi-head model is constructed with a specific emphasis on modularity, thereby enforcing a separation of concerns in design. 
The joint training of these components is carried out as a single integrated model, 
with the parameters denoted as $\theta = (\eta, \xi, \psi)$,
and the details described in the next section.

\vspace{.5ex}
\noindent \textbf{Learning Adaptive Behaviors.}
We solve the Optimization Problem \eqref{eq:policy_optimization},
considering its unconstrained relaxation:
\begin{gather}
\label{eq:lagrangian}
\min_{\lambda \geq 0} \,
\max_{\pi_\theta \in \Pi} \,
\mathcal{J}(\theta, \lambda) =
\min_{\lambda \geq 0} \,
\max_{\pi_\theta \in \Pi} \,
\mathcal{J}_{R}(\theta) - \lambda \, \mathcal{J}_{C}(\theta)
\end{gather}
where $\mathcal{J}$ is the Lagrangian, and 
$\lambda \geq 0$ is the Lagrange multiplier which acts as penalty term.
%
%
%
The two optimizations steps are interleaved till convergence, 
seeking for a saddle point of the original problem 
which is a feasible solution.


\noindent \textit{Policy Update.}
Considering the model-free setting due to the lack of knowledge of other agents,
the true return and cost distributions are induced by the policy rollouts and unknown.
We use a policy-gradient algorithm \cite{schulman2017proximal}, 
jointly optimizing an actor $\pi_{\theta}$ and a critic $v_{\zeta}$ models.
The critic simply regresses the value estimates $v_{target}(z_t)$, 
minimizing the loss:
\begin{gather}
    \mathcal{L}_v(\zeta) = 
    \mathbb{E}_{t} \big[ 
    (v_\zeta(z_t) - v_{target}(z_t)) ^ 2
    \big]
\end{gather}
The actor is updated 
by maximizing the following loss
\begin{gather}
    \mathcal{L}_{\pi}(\theta) = \mathcal{L}_{R}(\theta) - \lambda_k \mathcal{L}_{C}(\theta) + \beta \mathcal{L}_{ent}(\theta) 
\end{gather}
where 
$\lambda_k$ is the Lagrange multiplier introduced in Eq.~\eqref{eq:lagrangian}
at the $k$-th update,
$\mathcal{L}_R, \mathcal{L}_C$ denote the surrogate clipped losses 
for cumulative rewards and costs \cite{schulman2017proximal},
and $\mathcal{L}_{ent}$ denotes the entropy bonus for exploration.
We use generalized advantage estimation (GAE) \cite{schulman2015gae} to trade-off bias and variance 
in the advantage estimates $\hat{A}_{R, t}, \hat{A}_{C, t}$ for return and cost respectively. 
The surrogate clipped losses are defined as: 
\begin{gather}
\label{eq:l_clip}
\mathcal{L}_R(\theta) = \mathbb{E}_t \big[ 
    min(
        r_t(\theta) \hat{A}_{R, t},
        clip(
            r_t(\theta),
            1 - \epsilon,
            1 + \epsilon        
        ) \hat{A}_{R, t}
        )
\big] \nonumber \\
\mathcal{L}_C(\theta) = \mathbb{E}_t \big[ 
    min(
        r_t(\theta) \hat{A}_{C, t},
        clip(
            r_t(\theta),
            1 - \epsilon,
            1 + \epsilon        
        ) \hat{A}_{C, t}
        )
\big] \nonumber \\
r_t(\theta) = \frac{\pi_\theta(a_t | z_t)}{\pi_{\theta_{old}}(a_t | z_t)} \nonumber
\end{gather}
\noindent \textit{Lagrange Multiplier Update.}
The Lagrange multiplier plays as an adaptive penalty in the unconstrained problem, 
to make the infeasible solutions sub-optimal.
We update the Lagrangian multiplier with
the PID update rule \cite{stooke2020pid_lagrangian},
because of its effectiveness and simplicity in the implementation.
This update rule resembles the tuning of PID controllers
to correct oscillations and overshooting of traditional Lagrangian methods \cite{ray2019benchmarking}.
The update rule at iteration $k$ is as follows:
\begin{align}
\label{eq:pid_lagrangian}
    \lambda_{k} \leftarrow (K_P \Delta_k + K_I I_k + K_D \delta_k)_+
\end{align}
where $K_P, K_I, K_D \in \mathcal{R}_+$ are hyperparameters for the 
proportional, integral and derivative errors,
defined as:
\begin{gather}
    \Delta_k = \mathcal{J}_{C, k} - d \\
    I_k = (I_{k-1} + \Delta_k)_+ \\
    \delta_k = \mathcal{J}_{C, k} - \mathcal{J}_{C, k-1}
\end{gather}

\vspace{.5ex}
\noindent \textbf{Low-level Control Design.}
\label{section:control_design}
%
%
%
%
%
This section presents the design of CBF for multi-agent systems, which involves two steps:

\begin{enumerate}
    \item \textit{State and dynamics identification}: 
    we model the underlying system's dynamics.
    This step can follow first principles, employing physical laws and motion equations, 
    data-driven approaches, or a combination of these.
    
    \item \textit{CBF design}: 
    we design CBFs to enforce safety, 
    mapping states to numerical values. 
    This step presents several challenges
    in formalizing safety, defining barriers,
    efficient modeling and parameters selection to ensure feasibility while balancing safety and performance.
    
\end{enumerate}
To exemplify this methodology, 
we show the low-level control design of our motivating example.

\noindent \textit{Multi-robot system.}
Consider a system with $n$ agents \cite{borrmann2015swarm_cbf},
where each agent $i$ has non-linear control-affine dynamics:
\begin{gather}
    \label{dynamics:particle_env}
    s_{i, t+1} = 
    \left[
    \begin{matrix}
         p_{i, t+1} \\
         v_{i, t+1} \\
    \end{matrix}
    \right]
    =
    \left[
    \begin{matrix}
         I & \Delta t \\
         0 & I \\
    \end{matrix}
    \right]    
    \left[
    \begin{matrix}
         p_{i, t} \\
         v_{i, t} \\
    \end{matrix}
    \right]
    +
    \left[
    \begin{matrix}
         0 \\
         \Delta t \\
    \end{matrix}
    \right] a_{i, t}
\end{gather}
where $\Delta t$ is the discrete time-step, 
$p_i \in \mathcal{R}^2$, $v_i \in \mathcal{R}^2$, $a_i \in \mathcal{R}^2$ 
denote the position, velocity and acceleration of robot $i$ respectively.
We can write the joint multi-agent system as:
\begin{gather}
    \label{dynamics:joint_particle_env}
    s_{t+1} = 
    \left[
    \begin{aligned}
         p_{t+1} \\
         v_{t+1} \\
         z_{t+1} \\         
    \end{aligned}
    \right]
    =
    \left[
    \begin{aligned}
         f_p(s_t) \\
         f_v(s_t) \\
         f_z(s_t) \\         
    \end{aligned}
    \right]
    +
    \left[
    \begin{aligned}
         g_p(s_t) \\
         g_v(s_t) \\
         g_z(s_t) \\         
    \end{aligned}
    \right] a    
\end{gather}
where $p \in \mathcal{R}^2$ and $v \in \mathcal{R}^2$ denote the position and velocity of the ego agent,
$a \in \mathcal{R}^2$ denotes the ego action,
and $z \in \mathcal{R}^{n-4}$ denotes the other agent's states.
The real-valued functions $f_o, g_o$ are known for $o \in {p, v}$.
However the other agents' actions are unknown for the ego (i.e., $g_z=0$).
Without loss of generality, they can be assumed to be a function of the joint state $s_t$ 
and part of the dynamics $f_z$.
We consider the CBFs as pairwise safety constraints between the ego agent $i$ and any other agent $j \neq i$:
\begin{align}
    \label{cbf:particle_env}
    h(x) = 
    \frac{\Delta p_{ij}^{T}}{|| \Delta p_{ij} ||} \Delta v_{ij} 
    +
    \sqrt{
    a_{max} (|| \Delta p_{ij} || - D_s)
    }
\end{align}
where $a_{\text{max}}$ denotes the maximum braking that the ego agent can apply to avoid a collision, 
$\Delta p_{ij}$ represents the relative position $p_i - p_j$, and $\Delta v_{ij}$ the relative velocity $v_i - v_j$.



\noindent \textit{Multi-agent Racing.}
We consider a second use case of competitive multi-agent racing. 
The dynamics are modeled using an Euler-discretization of the kinematic bicycle model as in \cite{he2021rulecbf}.
We consider two safety specifications for collision with walls and opponents,
and model them using distance on Frenet and Cartesian coordinates, respectively.
For conciseness, we describe the dynamics and CBFs in the Appendix.

\section{Experiments}
\label{section:experiments}

In this section, we describe the experiments to evaluate our adaptive safe-learning approach in multi-agent systems.

\vspace{.5ex}
\noindent \textbf{Simulation.}
We conducted our experiments in the multi-agent environments presented in the previous section.
For the multi-robot system, we use the simulator and CBF from \cite{cheng2020safe} 
with its simplest collision-avoidance formulation.
For the multi-agent racing system, we use the F1tenth simulator \cite{okelly2020f1tenth},
which provides simulation of multiple vehicles and sensory inputs.
In both the environment, the CBF uses a constant-velocity model (CVM) 
for the other-agents behaviors.

\noindent \textbf{Training.}
We implemented the ASRL algorithm with the \texttt{omnisafe} library \cite{ji2023omnisafe}.
During training, we randomize the starting conditions and collect episodes of 15 seconds.
The agent observes the last $5$ states and learns with progress-based reward and sparse cost signals:
for the multi-robot system, the cumulative reward is $1$ for reaching the goal location and 
the cost is $1$ for collision with any opponent;
for the racing system, the reward is proportional to the relative distance in front of other vehicles and the cost is $1$ for collisions.
%
%
We evaluate the agent by averaging the reward and cost with a moving average over the last $100$ episodes and
train the agents for $1$ million steps.
More details on the environments and training are reported in the Appendix.

\noindent \textbf{Agents' Randomization:}
To create the condition for adaptiveness,
we randomize the number of agents and policies in each environment.
In the multi-robot system,
we select between $3$ and $7$ agents at each episode and 
randomize their policies through the safety distance $D_s$ used to avoid obstacles.
In the multi-agent racing system,
we simulate $2$ vehicles starting in front of the ego vehicle and 
tracking a reference line with velocity profile
randomly scaled by a factor
normally distributed with $\mu=0.60$ and $\sigma=0.05$.

\begin{figure}[h]
    \centering
    \includegraphics[width=.95\columnwidth]{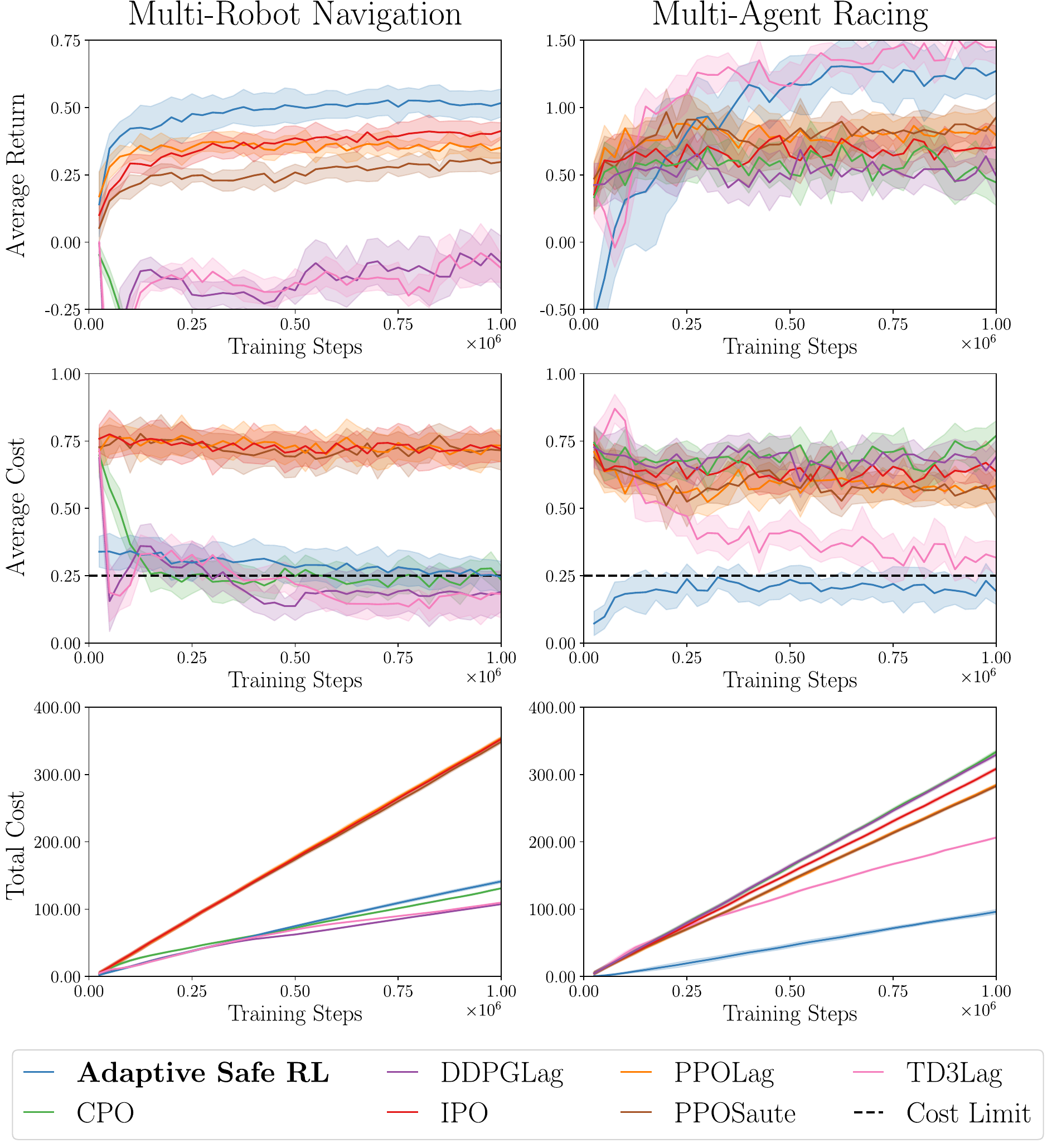}
    \caption{
    Learning curves of our approach and safe-RL baselines.
    Return, cost, and total cost averaged over $3$ runs.
    }
    \label{fig:learning_curves}
    \vspace{-4.5ex}
\end{figure}

\vspace{.5ex}
\noindent \textbf{Comparison End-to-End.}
In Figure \ref{fig:learning_curves}, we compare ASRL with the following 
state-of-the-art safe RL baselines:
PPO Lagrangian, DDPG Lagrangian and TD3 Lagrangian,
which are safe versions of the on-policy PPO \cite{schulman2017proximal}, and off-policy DDPG \cite{lillicrap2015ddpg} and TD3 \cite{fujimoto2018td3} respectively;
CPO, a trust-region method with near-satisfaction guarantees \cite{achiam2017cpo};
IPO, an interior-point policy optimization method \cite{liu2020ipo};
PPO Sauté, a state-augmentated PPO on the Sauté MDP \cite{sootla2022saute}.
%

In both environments, a noticeable performance gap emerges with on-policy algorithms
which struggle in achieving safe solutions within the desired limit. 
Their on-policy nature may require extended training periods to attain safe and optimal results.
Meanwhile, in the multi-robot environment,
the off-policy algorithms DDPG-Lag and TD3-Lag
achieve safe solutions but at the expense of poor overall performance.
Similarly, in the multi-agent racing environment,
on-policy algorithms continue to fail, while notably, 
TD3-Lag exhibits slow but consistent progress,
reaching a near-optimal level of performance by the end of training.

In contrast, our ASRL approach, incorporating trainable policies and CBF coefficients, 
fastly converges to high returns while consistently staying around the desired cost limit. 
Notably, in the multi-robot task, it starts above the cost limit and gradually reaches it, 
whereas in the racing task, it initiates below the limit and gradually adjusts safety levels to approach the threshold.
Our results suggest that our integration of CBF into the agent model 
improves exploration during training, 
resulting in a reduction in cumulative violation costs comparable to or better than off-policy methods, 
all while achieving significantly higher performance. 
This trend is graphically depicted Figure \ref{fig:learning_curves} (bottom).

\vspace{.5ex}
\noindent \textbf{Ablation Study on Learning Components.}
In this experiment, 
we evaluate the impact of adaptive safety 
and demonstrate its domain-adaptation skills.
To assess adaptive safety in isolation, 
we replace the policy module with a non-trainable controller.
During training, we use a Perturbed Gaussian policy \cite{rahman2022rpo} to foster exploration.
Details on training and controllers are reported in the supplementary material.

\begin{figure}[hb!]
   \centering
   \includegraphics[width=0.97\columnwidth]{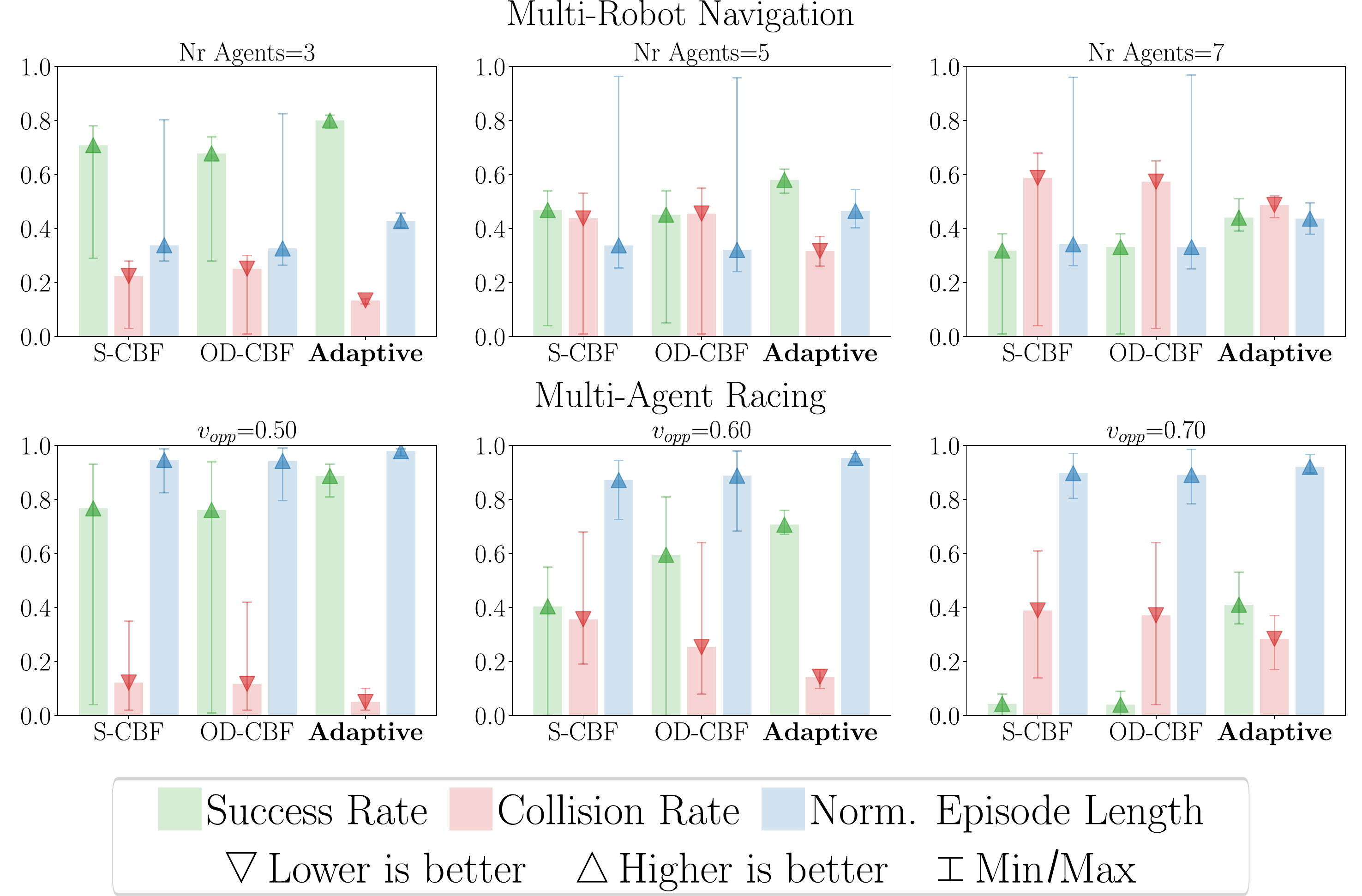}
   \caption{
   Comparison of the ablate model with control-theoretic approaches using the same nominal controller. 
   The bars show the mean rate with min/max delimiters for the same method.
   Performance for trained models are averaged over $3$ runs.
   }
   \label{fig:exp_fix_vs_adaptive}
   \vspace{-2.5ex}
\end{figure}

We evaluate the performance of the ablate model against traditional control-theoretic approaches:
Standard CBF (S-CBF), which uses CBF with fixed class-$\mathcal{K}$ coefficients,
and 
Optical-Decay Adaptive CBF (OD-CBF), which adapts the coefficients to ensure point-wise feasibility \cite{zeng2021safety}.
%
To account for the fact that these methods' performance are highly sensitive to the coefficients initialization,
we discretize the range of CBF parameters in $10$ values to cover most of the possible configurations.
We collect $100$ episodes for each configuration and report the performance in Figure \ref{fig:exp_fix_vs_adaptive}.
We observe comparable performance of OD-CBF and S-CBF in both the environments,
confirming that optimal-decay adaptiveness might improve feasibility of the QP problem but 
cannot capture long-term objectives, as sparse and delayed events of success or collision.
Conversely, our ablated model outperforms the baselines by simply adapting the CBF coefficients based on interactions with other agents.
This strongly suggests that learning of adaptive safety can substantially 
enhance the performance of existing controllers.

\vspace{-1.5ex}

\begin{figure}[hb!]
    \centering
    \includegraphics[width=.97\columnwidth]{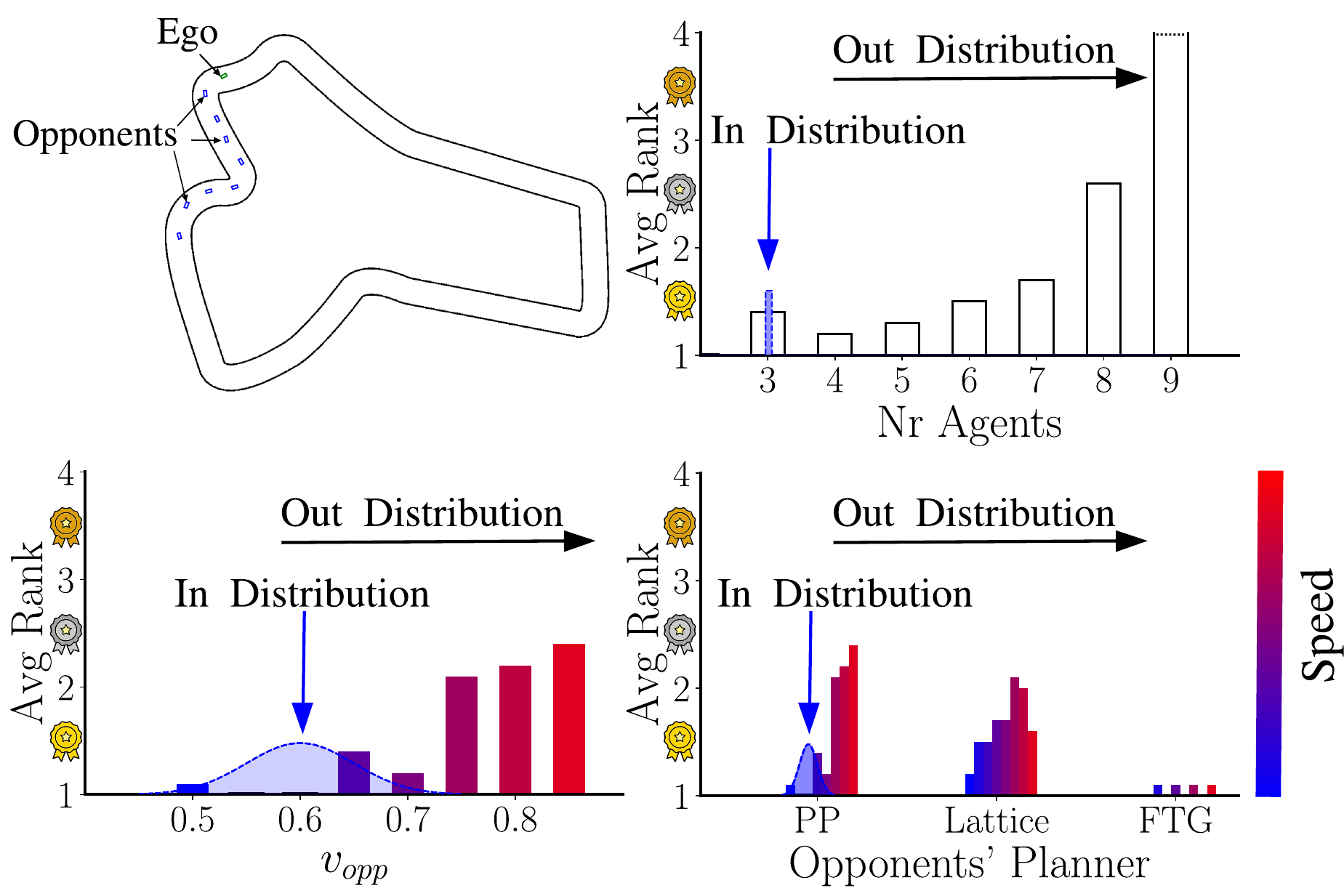}
    \caption{
    Generalization in multi-agent racing (\textit{top left}). 
    Performance measure the ego rank (\textit{lower is better}) 
    under previously unseen number of agents, velocity profiles $v_{opp}$, and planners.
    The training distribution is overlayed (\textit{blue}).
    }
    \label{fig:generalization}    
    \vspace{-2.25ex}
\end{figure}

\vspace{.5ex}
\noindent \textbf{Generalization and Scalability.} 
%
To evaluate the generalization capabilities of our trained agent in multi-agent systems,
we consider diverse racing scenarios including:
(1) varying the number of agents, 
(2) in-distribution planners with varying velocity profiles, and 
(3) out-of-distribution planners with new strategies and varying velocity profiles.
We focus on in- and out-distributions opponents, 
deliberately excluding cross-track generalization because competitive high-speed racing demands specialized strategies,
even to the extent of 
overfitting to track conditions.
For each scenario, 
we sample $10$ starting positions with the ego behind 
(Figure \ref{fig:generalization}, top left) and run simulations for $60$ seconds.
We consider collision or lap completion as termination conditions
and measure the final positioning (\textit{rank}) as common in racing competitions.

As shown in Figure \ref{fig:generalization},
%
the trained agent exhibits competitive performance,
often reaching 1st and 2nd place despite the initial positional disadvantage.
Training with $2$ opponents proves sufficient for generalization to races with many agents.
Up to $8$ opponents, the agent consistently secures a podium spot with an average rank below 3rd place.
However, with $9$ agents or more, 
the average rank exceeds the $4$-th position due to the limited time horizon for race completion.

For \textit{in-distribution} planners, 
the agent shows good adaptation and remains robust to faster velocity profile.
Notably, the rank increase appears directly linked to the training distribution.
For \textit{out-distribution} planners, 
we consider reactive (FTG~\cite{sezer2012ftg}) and sampling-based planners (Lattice~\cite{ferguson2008lattice}) with different velocities.
Our agent outperforms FTG, as expected due to its reactive nature which lacks of any global raceline. 
Moreover, we observe a competitive racing style against the Lattice planner, a robust baseline for comparison.
Notably, our agent maintains a high level of performance even against high-velocity profiles (\textit{right-most bars}),
suggesting the ability to learn characteristics behaviors for racing, and effectively reuse them against previously unseen opponents.

\vspace{-2ex}

\vspace{.5ex}
\noindent \textbf{Summary of Results.}
We assess our adaptive approach in two multi-agent systems with a range of diverse agents. 
The experiments revealed that our hierarchical integration of CBF
facilitated convergence to near-optimal agents,
outperforming a variety of safe RL baselines.
%
Moreover, our ablation study isolated the impact of Adaptive CBF,
showcasing superior adaptiveness than traditional control-theoretic methods. 
Finally, empirical evidence demonstrated our agent's ability to adapt and generalize across various racing scenarios, 
including unseen opponents and high-velocity profiles. 
%

\section{Related Work}

\noindent\textbf{Safe RL via CBF.}
Safe RL has drawn much attentions to prevent visiting unsafe states in safety-critical systems~\cite{garcia2015comprehensive, berkenkamp2017safe, chow2018lyapunov, alshiekh2018safe}.
The application of CBF in Safe RL has been proposed in \cite{cheng2019end} 
and is getting popular 
because of its safety guarantees and computational efficiency \cite{choi2020reinforcement, wang2023enforcing}.
Existing works train an RL agent to propose actions and use vanilla CBF to enforce safety.
However, to the best of our knowledge,
we are the first jointly training state-dependent CBF coefficients with an RL policy
to ensure bounded chance of violations.

\vspace{.5ex}
\noindent\textbf{Adaptive CBFs.}
Several works focus on improving feasibility and performance of CBF-controllers \cite{zeng2021safety, parwana2022recursive, xiao2023barriernet, xiao2021adaptive, parwana2022trust},
mostly focusing on single-agent or cooperative systems.
In these settings,
the CBF coefficients are optimized through gradient-based methods~\cite{parwana2022recursive}
or
policy distillation of a network with differentiable CBF layer,
under the assumption of available expert demonstrations~\cite{xiao2023barriernet}.
%
None of these works focus on multi-agent environments and adaptation to changing policies.
In contrast, our approach leverages discrete-time CBF, which better fits MDP theory, 
to train a model tailored for multi-agent adaptation with online RL.

CBF coefficients are updated in~\cite{parwana2022trust} based on the level of cooperation of other agents towards the ego.
However, they assume the ego agent knows the other agents' actions information beforehand.
Morevoer, they do not offer a direct way to update CBF coefficients but only mentioning that the derivative of them is monotonically increasing w.r.t. the level of cooperation, still requiring user intervention.
In contrast, we do not rely on such an assumption and leverage RL to train the coefficients, 
thus replacing any manual effort with a systematic methodology.
Moreover, we demonstrate our approach in a challenging multi-agent racing scenario.

\vspace{.5ex}
\noindent \textbf{Multi-agent CBF.}
In multi-agent systems, prior research proposed CBF to ensure collision-free behavior \cite{wang2017safety, borrmann2015swarm_cbf}.
Among these, 
~\cite{chen2020guaranteed} proposes a scalable decentralized approach to control multiple agents,
~\cite{cheng2020safe} presents a robust CBF with uncertainty model learned from data,
and \cite{qin2021learning} focuses on the joint optimization of control policy and CBFs.
However,
these works do not operate within an RL setting and do not
consider adaptation to many agents' policies.
Moreover, they mostly rely on fixed class-$\mathcal{K}$ function.


\section{Conclusions}
We present Adaptive Safe RL (ASRL) for multi-agent systems with partial observability from interactions with other agents.
Our novel ASRL combines model-free RL and adaptive CBF
to optimize long-term objectives under diverse agent strategies
while adhering the desired cost constraint.
ASRL surpasses traditional learning-based and control-theoretic approaches,
demonstrating adaptiveness and generalization across various multi-agent conditions, such as number of agents and their parameters.
Thus, ASRL enables safe autonomy in dynamic multi-agent setting.

\vspace*{1mm}\noindent\textbf{Why ASRL in multi-agent systems?}
CBFs are a valuable tool but their design and tuning is challenging with multiple agents.
We enhance CBFs with adaptive coefficients, 
integrating them into a trainable architecture and optimize them to diverse behaviors and long-term objectives.

\vspace*{1mm}\noindent \textbf{How does ASRL compare to existing approaches?}
ASRL retains benefits of CBF over learning methods, 
enabling efficient exploration while consistently adhering to cost limits.
Compared to control-theoretic methods, 
our trainable model achieves superior performance and adaptability.

\vspace*{1mm}\noindent \textbf{What are the limitations of ASRL?}
We primarily focus on systems with a relative degree of $1$.
To consider higher-order systems and CBFs~\cite{xiao2019control},
it would be possible to introduce multiple coefficients in our approach.
However, expanding the action space can make high-dimensional continuous control and exploration challenging. 
To address this, careful modeling of the action space is essential for tractability.
Also, 
ASRL does not assume full observability of other agents or explicit uncertainty quantification, thereby limiting its ability to guarantee safety in all scenarios.
While CBF typically assumes perfect knowledge of the system, this assumption rarely holds in practical scenarios. 
To this limitation, we adopt a chance-constraint formulation and ensure safety within certain bounds. 
Ongoing research is exploring alternative methods and uncertainty quantification in pursuit of robust solutions.

\section{Acknowledgements}
L.B. was supported by the Doctoral College in Resilient Embedded Systems (DCRES).


\appendix

\subsection{Nominal planners for Ablation study}
\label{appendix:nominal_planners}
In the ablation study, we train the adaptive safety module and control the ego with a built-in controller.
In this section, we describe the controllers adopted in each use case.

\vspace{.5ex}
\noindent \textit{Multi-Robot System.}
We use model predictive control (MPC)~\cite{garcia1989model} to generate (potentially unsafe) actions for the ego robot. 
The MPC steers the robot towards the goal and it is unaware of the other agents,
relying on the use of CBF for avoiding collisions.
Specifically, our nominal controller solves the following constrained finite horizon optimal control problem at each time step $t$:
\begin{subequations}
\label{eq:mpc-cbf}
\begin{align}
    a_{t:t{+}N{-}1}{=}&\argmin_{a_{t:t{+}N{-}1|t}} p(s_{t+N|t}){+}\sum_{k=0}^{N-1}q(s_{t+k|t}, a_{t+k|t}) \label{eq:mpc-cbf-cost}\\
    \text{s.t.} \quad 
    &s_{t+k+1|t} = f(s_{t+k|t}) + g(s_{t+k|t})a_{t+k|t},\label{eq:mpc-cbf-dynamics} \\
    &s_{t+k|t} \in S, a_{t+k|t} \in A \label{eq:mpc-cbf-constraint}\\
    &s_{t|t} = s_t, \label{eq:mpc-cbf-initial-condition}\\
    &s_{t+N|t} \in S_f, \label{eq:mpc-cbf-terminal-set}
\end{align}
\end{subequations}
where $N$ is the horizon, $p$ is the final state cost, $q$ is the state and control cost for each time step, $k$ is ranging from $0$ to $N-1$, (\ref{eq:mpc-cbf-dynamics}) is the joint dynamic of multi-agent system shown in~(\ref{dynamics:joint_particle_env}), and $S_f$ is the desired final state set.
Note that the system is evolved by applying $a_t$ at each time step, and then solve the above MPC again and obtain the next action to apply.
In our case, we choose 
$$p(s_t)=-s_t^TQs_{goal}$$ and 
$$q(s_t, a_t)=s_t^TQs_t+a_t^TRa_t,$$
where $s_{goal}$ is the goal state and 
\begin{gather}
    Q=
    \left[
    \begin{matrix}
         10 & 0 & 0 & 0 \\
         0 & 10 & 0 & 0 \\
         0 & 0 & 1 & 0 \\
         0 & 0 & 0 & 1 \\
    \end{matrix}
    \right],~
    R=0.02\,I_{4}.
\end{gather}


\vspace{.5ex}
\noindent \textit{Multi-Agent Racing.}
We use a sampling-based planner, known as lattice planner. 
It observes the current poses and velocities of the ego vehicle and two closest opponents,
and knows the raceline as a sequence of waypoints, generated as in~\cite{heilmeier2019minimum}.
Then, we sample local goals in a grid around the optimal raceline and generate corresponding dynamically feasible trajectories from the current pose to each local goal (see, e.g., ~\cite{mcnaughton2011motion, howard2009adaptive}). 
Each of the local plan is evaluated based on the following metrics: 
\begin{itemize}
    \item Cost based on the arc length, to prefer shorter trajectories;
    \item Cost based on maximum curvature, to avoid acute steering;
    \item Cost based on the similarity to the previous plan;
    \item Cost based on the tracking error to the raceline;
    \item Cost for collision with opponent and track boundary;
    \item Cost for low speed, to encourage fast racing;
    \item Cost for the co-occurrence of high speed and high curvature.
\end{itemize}

\subsection{Modeling of Multi-agent Racing System}
\label{appendix:racing_low_level}
The state of each racing agent $i$ at time $t+1$ is described by the following variables:

\begin{align}    
    \label{dynamics:kinematic_bicycle}
    \left[
    \begin{aligned}
         x_{i, t+1} \\
         y_{i, t+1} \\
         \psi_{i, t+1} \\         
         v_{i, t+1} \\
         e_{i, t+1} \\
         s_{i, t+1}
    \end{aligned}
    \right] 
    =
    \left[    
    \begin{aligned}[c]
         x_{i, t} &+ v_{i, t} \, cos (\psi_{i, t} + \beta_i) \, dt& \\
         y_{i, t} &+ v_{i, t} \, sin (\psi_{i, t} + \beta_i) \, dt& \\
         \psi_{i, t} &+ \frac{v_{i, t}}{l_r} sin \beta_i \, dt& \\
         v_{i, t} &+ a_i \, dt& \\
         s_{i, t} &+ \frac{v_{i, t} \, cos (\psi_{i, t} + \beta_i - \psi_{c}(s_{i, t}))}{1 - e_c \, k_c(s_{i, t})} \, dt& \\
         e_{i, t} &+ v_{i, t} \, sin (\psi_{i, t} + \beta_i - \psi_{c}(s_{i, t})) \, dt&
    \end{aligned}
    \right]
\end{align}

where for agent $i$ and time $t$:
\begin{itemize}
    \item $x_{i, t}, y_{i, t}, \psi_{i, t}$ represent the cartesian position and orientation,
    \item $v_{i, t}$ denotes the velocity,
    \item $e_{i, t}, s_{i, t}$ represent the lateral and longitudinal Frenet coordinates,
    \item $\beta$ denotes the slip angle, computed from the steering angle $\delta_i$ using the relation $$\beta_i = \tan^{-1} \left( \frac{l_r}{l_f + l_r} \tan (\delta) \right),$$
    for $l_r$ and $l_f$ representing the distances from the center of gravity to the rear and front axles, respectively.
\end{itemize}

The model also considers the heading of the track along the center-line, denoted by $\psi_c$, 
and the curvature of the center-line, denoted by $k_c$. 
These parameters are necessary to compute the evolution of $s_i, e_i$.
We rewrite the control inputs as $a_i, \beta_i$ instead of the 
usual $a_i, \delta_i$ to use a more tractable model with relative-degree $1$.

\paragraph{Control Barrier Function}
In the context of racing, 
we consider two different types of collisions,
respectively with walls and other vehicles.

\begin{itemize}
    \item To handle the collision with the wall, 
we use the Frenet coordinates and define a safety as keeping the lateral coordinate within a safe margin $d_{margin}$ from the track boundaries.
To do that, for each agent $i$ at time $t$,
we consider the lateral coordinate $e_{i,t}$, 
its velocity $\dot{e}_{i,t}$, 
and assume the agent can brake as $a_{\text{brake}}$. 
Formally, the CBF for wall collision is defined as:
\begin{gather}
    \label{cbf:f110_wall}
    h_{wall}(x) = 
    |e_{wall} - e_{i, t}| - \frac{\dot{e}_{i, t}^2}{a_{brake}} - d_{margin}.
\end{gather}

\item For collisions with other opponents, 
we use the same formulation as in the multi-robot system (e.g., with opponent $j$):
\begin{gather}
    \label{cbf:f110_opponent}    
    h_{opp}(x)\! =\! 
    \frac{\Delta p_{ij}^{T}}{|| \Delta p_{ij} ||} \Delta v_{ij} 
\!+\!
    \sqrt{
    a_{max}(v_{i, t}) (|| \Delta p_{ij} || \!-\! D_s)
    }.
\end{gather}
\end{itemize}

However, 
we introduce a velocity-dependent braking acceleration 
$a_{\text{max}} = a_{max} \left( 1 - \frac{v_{i, t}}{v_{max}} \right)$
to discourage driving at high speeds in the direction of an opponent, as it would increase the chance of collision.

\subsection{Environments}
\label{appendix:environments}

In this section, we describe the details on the observations, actions, and training signals 
used in each environment.

\noindent \textit{Multi-Robot System.}
The agents positions and controllers are randomized at the beginning of each episode.
All non-controllable agents drive towards their goal avoiding obstacles within a distance $D_s$,
thus sampling $D_s$ for each agent give a diverse set of policies to cope with.
As an example, an agent with $D_s=0$ would result in aggressive maneuvers to reach the goal position as soon as possible,
while $D_s=6$ would proceed while avoiding other agents closer than $6$ units from it.
In our experiments, we sample $D_s$ among $0, 5, 6$ with about $50\%$ chance of $0$
and $50\%$ chance of $5$ or $6$.

The ego agent observes the states of the $k$ closest agents,
including their position $x, y$, velocity $v_x, v_y$ and goal $x_g, y_g$.
We do not expose the policy parametrization of other agents.

The ego actions are intermediate waypoints $x_{wp}, y_{wp}$ relative to the ego coordinate
and bounded within $1.0$ unit from it. 
The min-time controller presented in the previous section is used to compute the accelerations to reach the goal.

We use a progress-based reward defined as
$$
r(s, a, s') = \frac{d_{eu}(s) - d_{eu}(s')}{d_{eu,0}} \\
$$
where $d$ denotes the Euclidean distance with the goal position,
and $d_0$ is the distance from the initial state, used as a normalization constant .
The cost builds on a sparse signal in case of the collision event, 
which is $1$ if the ego agent is in collision with any other agent.
In case of collision, the episode terminates.

\noindent \textit{Multi-Agent Racing.}
The vehicles positions are randomized at the beginning of each episode,
sampling a position in the first-half of the track.
The non-controllable agents track a reference trajectory, precomputed offline as in \cite{heilmeier2019minimum}.
We vary the agents behavior by scaling the velocity profile of the reference trajectory by a random factor around $0.6$.

The ego vehicle starts at the end of the batch of vehicles and drives without any velocity scaling,
so that it can catch up with the other vehicles and overtake them.
In the observation, we include the agents poses $x, y, \theta$ and positions in Frenet coordinates $s, e$ with respect to centerline,
and some track features, including $10$ raceline waypoints with curvature for a lookahead distance of $10$ meters.

The actions consist of local plans as cubic splines. 
We characterized them in Frenet frame, by controlling the lateral displacement $e_f$ and target velocity $v_f$.
From the current position of the vehicle, we use an adaptive lookahead distance based on the velocity
$$
l(v) = l_{min} + \frac{v * (l_{max} - l_{min})}{l_{scale}}
$$
and derive the target waypoint corresponding to $l(v), e$.
Then, we fit a cubic spline from the current position to the target waypoint and use the target velocity as reference.

We reward the agent based on the relative progress in front of the other agents at the end of the episode,
saturated to $+1$ when at least $5$ meters ahead and $-1$ when at least $5$ meters behind.
In particular, we use the following signal:
$$
r(s, a, s') = \sum_{agent \neq ego} tanh \left( \frac{d_{fr}(s_{ego}, s_{agent})}{5.0} \right)
$$
where $d_{fr}$ denotes the distance in the longitudinal Frenet coordinate,
and $5.0$ serves as a coefficient to account for the car length and cap the rewarding beyond a 
sufficient margin to be considered as overtaking.

The cost builds on a sparse signal in case of the collision with the wall or any opponent, 
which is $1$ if the ego agent crashes into them. Even in this case, when the ego collides at fault or not, the episode terminates.
We do not terminate the episode if the other agents collide among them.

\subsection{Training Details}
\label{appendix:training}

In this section, we discuss the agent training and the specific settings we used. Table \ref{tab:hyperparams} provides detailed information on these settings for each experiment.
In our Ablation Study, we identified two key decisions that significantly impact the performance:
(1) changing the actor distribution to Perturbed Gaussian;
(2) reducing the frame-stacking to single observation.
We observe the ablation study to make the learning process harder because the degree of freedom of the agent are restricted to the adaptive safety only, 
due to the use of built-in controllers. The resulting reward and cost become noisy and a training signal difficult to extract.
For this reason, we introduce Perturbed Gaussian to help in exploration and avoid the early convergence to sub-optimal distributions in favour of high-entropy ones.
Moreover, smaller frame stacking significantly reduces the observation dimensions, 
especially in the racing task, where the observations include many redundant track features.
In future work, this approach could benefit from better model architectures to capture the task features in a more effective way.

\begin{table*}[h]
  \centering
  \begin{tabular}{|ccccc|}
    \hline
    \textbf{Parameter} & \textbf{End-to-End} & \textbf{End-to-End} & \textbf{Ablation Study} & \textbf{Ablation Study}\\    
    \textbf{Name} & \textbf{Multi-Robot Nav.} & \textbf{Multi-Agent Racing} & \textbf{Multi-Robot Nav.} & \textbf{Multi-Agent Racing}  \\
    \hline
    Evaluation frequency & 10 & 10 & 10 & 10 \\
    Nr Evaluation Episodes & 20 & 20 & 20 & 20 \\
    Total Timesteps & 1M & 1M  & 1M & 1M \\
    Learning Rate & 3e-4 & 3e-4  &3e-4 & 3e-4\\
    Nr Environments & 4 & 2  & 4 & 2 \\
    Nr Steps & 2048 & 2048  & 2048 & 2048 \\
    LR Annealing & True & True  & True & True \\
    Discount Factor & 0.99 & 0.99  & 0.99 & 0.99 \\
    GAE $\lambda$ & 0.95 & 0.95  & 0.95 & 0.95 \\
    Nr Minibatches & 32 & 32  & 32 & 32 \\
    Update Epochs & 10 & 10  & 10 & 10 \\
    Normalize Advantages & True & True  & True & True  \\
    Normalize Costs & True & True  & True & True  \\
    Clip Coeff. & 0.2 & 0.2  & 0.2 & 0.2 \\
    Clip Value Loss & True & True  & True & True  \\
    Entropy Bonus & 0.0 & 0.0  & 0.0 & 0.0 \\
    Value Function Coeff. & 0.5 & 0.5  & 0.5 & 0.5 \\
    Max Grad Norm & 0.5 & 0.5  &  0.5 & 0.5 \\
    Lagrange Multiplier Init & 1.0 & 1.0  & 1.0 & 1.0 \\
    Lagrange Multiplier Max & 100 & 100  & 100 & 100 \\
    Use-pid-lagrangian & True & True  & True & True  \\
    PID Lagrangian Gains & [10.0, 0.1, 0.1] & [1.0, 0.1, 0.1]  & [10.0, 0.1, 0.1] & [1.0, 0.1, 0.1] \\
    PID Lagrangian EMA Alpha & 0.95 & 0.95  & 0.95 & 0.95 \\
    PID Lagrangian Delay & 1 & 1   & 1 & 1 \\
    Cost Limit & 0.25 & 0.25   & 0.25 & 0.25 \\
    Frame Stack & 5 & 5  & 1 & 1 \\
    \hline
    \textbf{Agent} & & \\
    \hline    
    Actor Architecture & [64, 64] & [64, 64] & [64, 64] & [64, 64] \\
    Actor Activation & Tanh & Tanh & Tanh & Tanh \\
    Actor Distribution & Gaussian & Gaussian & Perturbed Gaussian ($\alpha=0.5$) & Perturbed Gaussian ($\alpha=0.5$) \\
    Critic Architecture & [64, 64] & [64, 64]  & [64, 64] & [64, 64] \\
    Critic Activation & Tanh & Tanh & Tanh & Tanh \\
    \hline
  \end{tabular}
  \caption{Experiments Hyperparameters.}
  \label{tab:hyperparams}
\end{table*}

\bibliographystyle{unsrt}
\bibliography{biblio}

\end{document}